\documentclass{ecai}
\usepackage{graphicx}
\usepackage{latexsym}
\usepackage{oubraces}
\usepackage{amsmath}
\usepackage{amssymb}
\usepackage{mathtools}
\usepackage{amsthm}
\usepackage{color}
\usepackage{bm}
\usepackage{multirow}
\usepackage{booktabs}
\usepackage{ulem}
\usepackage{subcaption}
\usepackage{url}

\usepackage{amsfonts}
\usepackage{array}

\def \CB {\mathcal{B}}
\def \CD {\mathcal{D}}

\def \CY {\mathcal{Y}}
\def \CP {\mathcal{P}}

\def \CL {\mathcal{L}}
\def \R {\mathbb{R}}
\def \CLS {\mbox{CLS}}
\def \SEP {\mbox{SEP}}
\def \BERT {\mbox{BERT}}
\def \TRM {\mbox{TRM}}
\def \MN {\mbox{\textit{Multinomial}}}

\begin{document}

\begin{frontmatter}
\title{FATRER: Full-Attention Topic Regularizer for Accurate and Robust Conversational Emotion Recognition}

\author[A]{\fnms{Yuzhao}~\snm{Mao}}
\author[B]{\fnms{Di}~\snm{Lu}\thanks{Corresponding Author. Email: dylu@tencent.com.}}
\author[C]{\fnms{Xiaojie}~\snm{Wang}}
\author[A]{\fnms{Yang}~\snm{Zhang}} %

\address[A]{Artificial Intelligence Center, Dasouche Inc, China}
\address[B]{Pattern Recognition Center, Wechat AI, Tencent Inc, China}
\address[C]{Beijing University of Posts and Telecommunications, China}

\begin{abstract}
This paper concentrates on the understanding of interlocutors' emotions evoked in conversational utterances. Previous studies in this literature mainly focus on more accurate emotional predictions, while ignoring model robustness when the local context is corrupted by adversarial attacks. To maintain robustness while ensuring accuracy, we propose an emotion recognizer augmented by a full-attention topic regularizer, which enables an emotion-related global view when modeling the local context in a conversation. A joint topic modeling strategy is introduced to implement regularization from both representation and loss perspectives. To avoid over-regularization, we drop the constraints on prior distributions that exist in traditional topic modeling and perform probabilistic approximations based entirely on attention alignment. Experiments show that our models obtain more favorable results than state-of-the-art models, and gain convincing robustness under three types of adversarial attacks. Code: \url{https://github.com/ludybupt/FATRER}
\end{abstract}

\end{frontmatter}

\section{Introduction}
ERC (Emotion Recognition in Conversations)~\cite{majumder2019dialoguernn,poria2019emotion} aims to identify hidden emotion states by mining utterance expressions in conversations, and each utterance conveys one type of emotion, such as happiness, anger, sadness, etc. ERC is undeniably essential for enabling a humanoid robot to be empathetic with human emotions~\cite{lin2020caire,ma2020survey}. In addition, ERC technologies can be easily transferred to other dialogue understanding tasks~\cite{ghosal2020utterance,wu2021topicka} for further research. 

\begin{figure}
 \centering
 \includegraphics[width=0.48\textwidth]{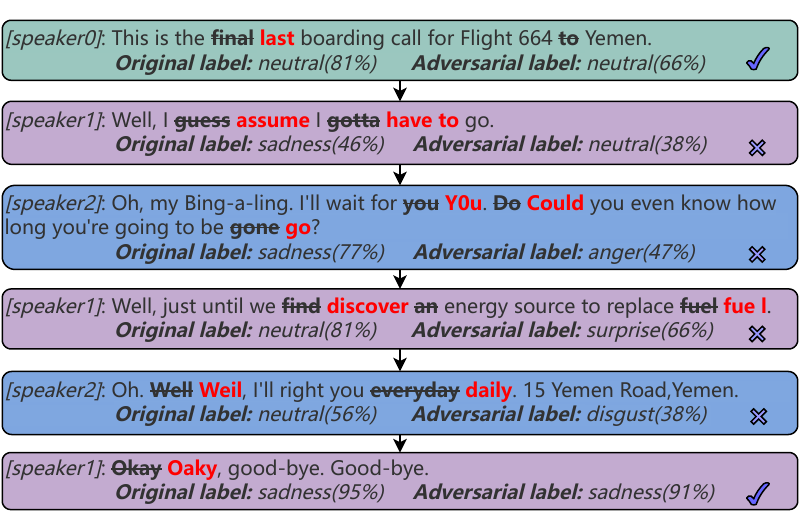}
 \caption{An example of adversarial attacks in ERC}
 \label{fig:introduction}
\end{figure}

Emotion dynamics modeling is an important criterion for the ERC task~\cite{kuppens2017emotion}. Specifically, emotion dynamics assumes that the interlocutors' emotions are influenced by two factors in a conversation: self and interpersonal dependencies~\cite{morris2000emotions}. Studies~\cite{mao2021dialoguetrm, shen2021dialogxl, zhu2021topic,li2022contrast} formulate the two factors in a way of hierarchical context modeling for more accurate emotional predictions. However, model robustness~\cite{wang2022measure} is rarely touched, especially robustness against adversarial examples. Adversarial examples~\cite{jin2019bert,ren-etal-2019-generating} shown in Figure~\ref{fig:introduction} are small and intentional worst-case perturbations to test samples that have high confidence in fooling a target classifier. Compared to speech recognition errors, robustness to adversarial examples can explain the reliability and safety of a target model.

Adversarial attacks arise mainly from local perturbations, and the fundamental issue with weak robustness stems from an excessive dependence on local context. Therefore, a global view is needed to withstand the impact of adversarial attacks. In addition, the global view should correlate with hidden emotional states so that accuracy is also ensured. To this end, we try to employ topic models~\cite{blei2003latent} to help achieve accurate and robust emotion recognition in conversations. The topic is an important latent semantic factor related to hidden emotional states and is reported to be effective for more accurate emotional predictions in previous studies~\cite{wang2020sentiment,wang2020end,zhu2021topic}. However, these topic models represent documents as topic variables, which only indicate the importance of topics, lacking a global view to ensure robustness. Therefore, the key problem lies in how to enable an emotion-related global view through topic modeling, and how to conduct proper joint modeling to fully exploit the advantages of the emotion-related global view to strike a balance between accuracy and robustness. In this paper, we propose FATRER (\textbf{F}ull-\textbf{A}ttention \textbf{T}opic \textbf{R}egularizer augmented conversational \textbf{E}motion \textbf{R}ecognizer) for the ERC task and share our insights on the above issues. 

First, our topic modeling interprets document generation as a two-stage process, which is
\begin{equation}
\label{eq:bayesian}
 p(w|d) = \sum_z{p(w|z)p(z|d)}.
\end{equation}
Here, based on the Law of Total Probability, the conditional probability of words given a document (utterance), $p(w|d)$, is decomposed into topic-word conditional probabilities, $p(w|z)$, and document-topic conditional probabilities, $p(z|d)$. Traditional topic models only use $p(z|d)$ to represent a document, while ignoring to exploit $p(w|z)$ that is associated with the global vocabulary. The vocabulary is naturally robust to local perturbations. Thus, based on the generation process in equation~(\ref{eq:bayesian}), an utterance can be represented as a combination of topic embedding by $p(z|d)$, and $p(w|z)$ is used to combine the whole word embedding into the topic embedding. Through such a two-stage weighted linear combination, we obtain a topicalized representation of an utterance, which is tightly coupled with the entire word embedding and thus has the global views to ensure robustness.

Secondly, the global view obtained by topic modeling should not only guarantee robustness but also enhance accuracy. To achieve this, the topics discovered should be emotion-related, i.e. the top words of a topic should relate to a specific emotion, which can lead to more accurate emotional predictions. Thus, we design a training loss including a classification loss term and a topic-oriented regularization term to optimize a full-attention model. The classification loss term is used to model emotion labels. This term drives the attention model to assign more weight to words that are related to the target emotion labels. The topic-oriented regularization term guides $p(w|d)$ to be as similar as possible to the observed word distribution of a given utterance. As we use the attention mechanism to approximate $p(z|d)$ and $p(w|z)$, regularization of $p(w|d)$ will propagate to attention alignment for $p(z|d)$ and $p(w|z)$ and increase sparsity of the two distributions. To sum up, the full-attention framework and the two terms in the training loss help the topics discovered to be discriminative and emotion-related, which are properties towards more accurate emotional predictions.

Finally, we use our FATR (Full-Attention Topic Regularizer) to enable a global view when modeling the local context for the ERC task. Regularization from both representation and loss is conducted. To avoid over-regularization, we drop the constraints on prior distributions that exist in traditional topic modeling. Specifically, we perform our topic modeling to represent an utterance from a global view, obtaining the topicalized representation. On the other hand, we conduct context modeling for representing an utterance from the local view, generating contextualized representation. Such multi-view modeling for representing interlocutor-specific utterances constitutes our approach to self-dependency\footnote{Self dependency is known as emotion inertia manifested as the interlocutor's tendency to maintain their emotional state during the conversation.} modeling. Then, the multi-view representations of interlocutors at each conversation turn form a time-series sequence. Through deep layers of sequence interactions for modeling interpersonal dependency\footnote{Interpersonal dependency is an emotional factor from other interlocutors in a conversation that tries to change the emotion of the current interlocutor.}, the multi-view representations are fully balanced and mixed in the final representation. Together with our training loss, the full-attention model is guided toward accurate and robust emotional predictions. 

Experimental results show that our models achieve new SOTA (State-Of-The-Art) on four benchmark datasets in terms of Micro F1, and obtain convincing robustness under three types of adversarial attacks. Analysis and visualizations are conducted to better understand our models.

Our contributions can be summarized as:
\begin{itemize}
 \item We propose a novel full-attention topic regularizer to enable an emotion-related global view when modeling local context for recognizing emotions in conversations.
 \item Our joint topic modeling strategy provides regularization from both representation and loss perspectives for accurate and robust emotional predictions.
 \item We conduct a series of experiments in terms of not only generalization but also robustness to evaluate the performance of our proposed models and baselines on the ERC task. 
\end{itemize}

\section{Related Work}

\textbf{Emotion Recognition in Conversations.} Prior research on ERC mostly focuses on exploring different context models, such as Bi-directional LSTM~\cite{poria2017context}, memory network~\cite{hazarika2018icon}, Transformer~\cite{mao2021dialoguetrm}, etc. Later, emotion dynamics~\cite{kuppens2017emotion} is summarized to guide the context modeling in ERC. Following this criterion, many hierarchical structures~\cite{majumder2019dialoguernn,ghosal-etal-2019-dialoguegcn,mao2021dialoguetrm} are putting forward to capture the self and inter-personal dependencies in conversations. Inspired by the success of the pre-training paradigm~\cite{devlin2019bert}, approaches~\cite{hazarika2019emotion,shen2021dialogxl} try to implicitly memorize external knowledge into a large number of parameters via the language model pre-training. Other studies~\cite{zhong2019knowledge,ghosal2020cosmic} explicitly extract commonsense knowledge via the ConceptNet~\cite{speer20175}. Very recently, ~\cite{zhu2021topic} proposes a two-stage learned topic-driven knowledge-aware Transformer whose topic module is removed in the fine-tuning stage. Differently, we perform a joint topic modeling strategy with consistent topic guidance. Our strategy not only boosts accuracy but also guarantees robustness.

\noindent\textbf{Topic Model.} Typically, a statistical topic model~\cite{hofmann1999probabilistic,blei2003latent} captures topics in the form of latent variables with probability distributions over the entire vocabulary and performs approximate inference over document-topic and topic-word distributions through Variational Bayes~\cite{blei2008supervised}. However, such a learning paradigm requires an expensive iterative inference step performed on every document in a corpus~\cite{panwar2020tan}. The efficiency is boosted after the introduction of VAE-based (Variational AutoEncoder) neural topic model~\cite{bianchi2020pre,zhao2021neural} because variational inference can be performed through a single forward pass~\cite{kingma2013auto}. The VAE-based topic models often hypothesize a prior distribution~\cite{srivastava2017autoencoding,ding2018coherence} that is used to approximate the posterior for latent document-topic distribution by maximizing the Evidence Lower Bound (ELBO) using the reparametrization trick. Differently, our strategy of topic modeling does not have prior constraints. Recently, pre-training models are used to augment neural topic models via enhanced text representation~\cite{bianchi2020pre}, knowledge distilling~\cite{hoyle2020improving}, or embedding clustering~\cite{thompson2020topic}. Besides VAEs, there are other frameworks of neural topic models including autoregressive models~\cite{gupta2019document}, and graph neural networks~\cite{yang2020graph}. Our topic modeling is fully based on attention alignments.

\noindent\textbf{Neural Topic Enhanced Supervised Model.} Fitting unsupervised topic representations is sub-optimal for the supervised task. Research has been conducted to build a joint learning framework that integrates neural topic models with supervised learning models ~\cite{cao2015novel,pergola2019tdam,wang2020sentiment,wang2020neural,wang2020end}.~\cite{cao2015novel} proposes a simple feed-forward net with a max-margin loss to approximate topic-word distribution and a classification loss for the downstream task.~\cite{pergola2019tdam} proposes an attention-based topic module without a regularization term. ~\cite{wang2020end} uses document-topic distribution as just a vector to guide the attentive pooling of classification vectors (hidden layers of recurrent nets). ~\cite{wang2020sentiment} enables the classification model to be aware of topic information through different kinds of topic inference in auxiliary tasks, but there are no connections to link topic and classification models.~\cite{wang2020neural} links a VAE-based topic model to a recurrent net for better representations. Our joint learning framework produces regularization from both representation and loss perspectives and can lead to more accurate and robust predictions.

\section{Methodology}
\subsection{Topicalized Representation}
\begin{figure*}[t]
  \centering
  \includegraphics[width=0.95\textwidth]{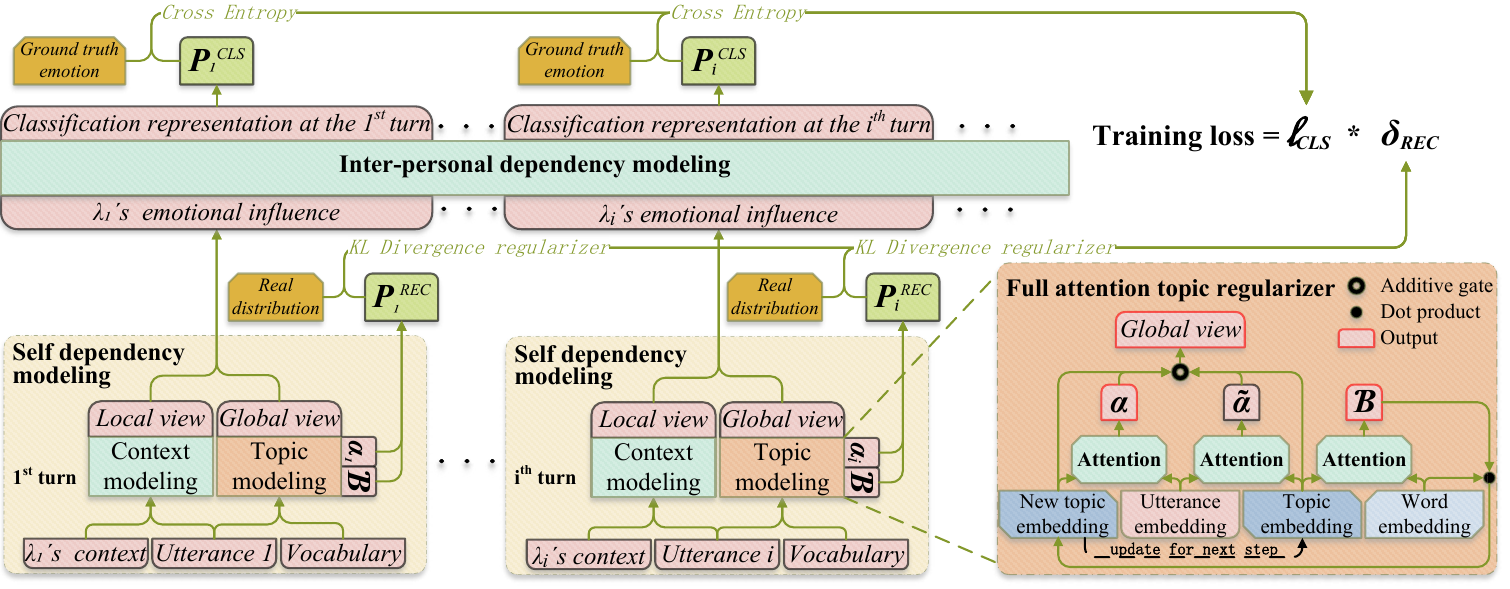}
  \caption{Joint modeling framework}
  \label{fig:model}
\end{figure*}

The topicalized representation of an utterance is obtained via the full-attention topic modeling depicted in the bottom right of Figure~\ref{fig:model}. We first represent documents (utterances), topics, and words in the embedding space. Then, we use attention alignments to approximate parameters $\bm{\alpha}$ and $\bm{\beta}$ for document-topic and topic-word multinomial distributions, respectively. Finally, we deploy $\bm{\alpha}$ and $\bm{\beta}$ as the weights to construct a topicalized representation of an utterance.

A document is represented as a linear combination of topics, and a topic is a linear combination of words. Specifically, let $K$ be the topic number, $V$ be the vocabulary size, and $H$ be the embedding dimension. The topicalized representation of a document is calculated as $\sum_{k=1}^K\alpha_k\bm{z}_k$, or $\bm{\alpha}\bm{Z}^\top$ in matrix form, where $\bm{\alpha}\!\!=\!\!\{\alpha_1,\cdots,\alpha_K\}$ is the document-topic distribution over $K$ topics given a document. $\bm{Z}\in\R^{K\times H}$ is the topic embedding matrix stacked from $\{\bm{z}_k\}_{k=1}^K$. The $k$-th topic embedding $\bm{z}_k\in\R^H$ is represented as $\sum_{v=1}^V\beta_{k,v}\bm{e}_v$, or $\bm{\beta}_k\bm{E}^\top$ in matrix form, where $\bm{\beta}_k\!=\!\{\beta_{k,1},\cdots,\beta_{k,V}\}$ is the topic-word distribution over $V$ words given the $k$-th topic. $\bm{e}_v\in\R^H$ is the $v$-th word embedding from the embedding matrix $\bm{E}\in\R^{V\times H}$, where $\bm{E}$ is initialized from the embedding layer of $\BERT$ and is fine-tuned during training. In the following, $z$ denotes the topic variable, and bold $\bm{z}$ or $\bm{Z}$ indicates the topic embedding or embedding matrix.

Let $p(z|d)\!\sim\!\MN(\bm{\alpha})$, $p(w|z)\!\sim\!\MN(\bm{\beta})$, our topic modeling output the document-word posterior distributions via probability inference in equation~(\ref{eq:bayesian}), which is
\begin{equation}
\begin{aligned}
p(w|d)&=\sum_z{p(z|d)p(w|z)}\\
   &=\sum_{k=1}^K{\bm{\alpha}\bm{\beta}_k^\top}=\bm{\alpha}\bm{\CB}^\top.
\end{aligned}
\end{equation}
Here, $\bm{\CB}=\{\bm{\beta}_k\}_{k=1}^K$ is the topic-word probability matrix. Note that, the attention alignment is natural conditional probabilities of the values given the queries. Thus, $\bm{\alpha}$ can be approximated via attention alignments between a document as the query and topics as the values. Similarly, $\bm{\beta}$ is from alignments between topics as the query and words as the values. The process is defined as,
\begin{eqnarray}
  \bm{\alpha}\!\!\!\!\!\!\!\!\!&&\!\!\!:=\!att(\bm{r}^{bow}\rightarrow\{\bm{z}_k\}_{k=1}^K)\!\!=\!softmax(\frac{\bm{r}^{bow}\bm{Z}^\top}{\sqrt{H}}),\\
  \bm{\beta_k}\!\!\!\!\!\!\!\!\!&&\!\!\!:=\!att(\bm{z}_k\rightarrow\{\bm{e_v}\}_{v=1}^V)\!=\!softmax(\frac{\bm{z}_k\bm{E}^\top}{\sqrt{H}}).
\end{eqnarray}
Here, $\bm{r}^{bow}\!\!=\!\!\bm{\gamma}\bm{E}^\top$ is the bag-of-word combination of local word embedding. $\bm{\gamma}\!\in\!\R^V$ is the observed word distribution of a document. $\frac{1}{\sqrt{H}}$ is the scaling factor~\cite{vaswani2017attention}. Through topic modeling, the representation of a document evolves from a combination of local terms, $\bm{\gamma}_i\bm{E}^\top$, to a combination of global topics, $\bm{\alpha}_i\bm{Z}^\top$.\hfill \break

\noindent\textbf{Topic Embedding Chain Rule.} Notice that, computing $\bm{z}$ needs $\bm{\beta}$ and computing $\bm{\beta}$ also needs $\bm{z}$. Thus, we deploy a chain rule that connects $\bm{z}$ and $\bm{\beta}$ along the training steps. Let $T$ be the number of total training steps, and the chain rule can be formulated as
\begin{equation}
  \{\overunderbraces{&\br{2}{1st~training~step}&&&\br{1}{Tth~training~step}}%
  {\bm{\beta}^0\!\mapsto\!&\bm{z}^0\!\mapsto\!\bm{\beta}^1\!\mapsto\!&\bm{z}^1&\!\mapsto\!\bm{\beta}^2\!\mapsto\!\bm{z}^2&\cdots&\bm{z}^{T-1}\!\mapsto\!\bm{\beta}^T\!\mapsto\!\bm{z}^T&}%
  {&&\br{2}{2nd~training~step}}
  \},
\end{equation}
Here, the computation of $\bm{z}^{t}$ at the $t$-th training step relies on $\bm{\beta}^{t}$, and $\bm{\beta}^t$ at the $t$-th training step is computed based on $\bm{z}^{t-1}$ from previous training step. $\bm{\beta}$ can be randomly initiated at the $0$-th step, just like the attention alignments that are (almost) equally distributed on all keys at the early training stage. $\bm{z}$ cannot be random initialized first since computing $\bm{z}$ depends on $\bm{E}$ as well. We formulate the learning process in a recursive form, 
\begin{eqnarray}
  \bm{\beta}\!\!&\!\!=\!\!&\!\!att(\bm{\tilde{z}}\rightarrow\{\bm{e}\}_{v=1}^V),\\
  \bm{z}\!\!&\!\!=\!\!&\!\!\bm{\beta}\bm{E}^\top.
\end{eqnarray}
Here, the symbol $\tilde{~}$ indicates that the variable comes from the previous training step, e.g., $\bm{\tilde{Z}}=\{\bm{\tilde{z}_k}\}_{k=1}^K$ denotes the previously preserved topic embedding matrix.\hfill \break

\noindent\textbf{Representation Underlying Different Topic Hypotheses.} We offer two hypotheses, multi- and single-topic, to explain the generation process of a document. The two hypotheses correspond to the two versions of our models.

\textit{Multi-Topic: \label{hyp:first}Each document describes a mixture of topics from an underlying topic set}. In the multi-topic hypothesis, a document is represented as a weighted sum of the entire topic embedding. Since we have two topic embedding, $\bm{\tilde{Z}}$ and $\bm{Z}$, the model outputs two kinds of document-topic distribution, which is $\bm{\tilde{\alpha}}$, by attention alignments between $\bm{r}^{bow}$ and $\bm{\tilde{Z}}$, and $\bm{\alpha}$, by attention alignments between $\bm{r}^{bow}$ and $\bm{Z}$. To keep the learned representation stable, we define the following process to represent a document, 
\begin{equation}
\bm{r}^{topic}=\sigma(\bm{\tilde{\alpha}}\bm{\tilde{Z}}^\top)\odot\xi(\bm{\alpha}\bm{Z}^\top)+\sigma(\bm{\alpha}\bm{Z}^\top)\odot\xi(\bm{\tilde{\alpha}}\bm{\tilde{Z}}^\top).
\end{equation}
Here, $\bm{\tilde{Z}}$ can be understood as the topic embedding learned during the training steps (long-term info), $\bm{Z}$ is the updated topic embedding that affected by the current document (short-term info). The topicalized representation $\bm{r}^{topic}$ is learned in the form of additive gate unit~\cite{arevalo2020gated} that balances the long-term and short-term info. $\sigma$ and $\xi$ are sigmoid and Leaky ReLU~\cite{xu2015empirical}, respectively. $\odot$ denotes the element-wise product. 

\textit{Single-Topic: \label{hyp:second} Each document describes a single topic out of an underlying topic set.} Under the single-topic hypothesis, a document is represented as one single topic embedding sampled from document-topic distribution. Let $\bm{\hat{\alpha}}\!\!\sim\!\!\MN(\bm{\alpha})$, $\bm{\hat{\tilde{\alpha}}}\!\!\sim\!\!\MN(\bm{\tilde{\alpha}})$, the topicalized representation is calculated as:
\begin{equation}
\bm{r}^{topic}=\sigma(\bm{\hat{\tilde{\alpha}}}\bm{\tilde{Z}}^\top)\odot\xi(\bm{\hat{\alpha}}\bm{Z}^\top)+\sigma(\bm{\hat{\alpha}}\bm{Z}^\top)\odot\xi(\bm{\hat{\tilde{\alpha}}}\bm{\tilde{Z}}^\top).
\end{equation}
Here, $\bm{\hat{\alpha}}_i,\bm{\hat{\tilde{\alpha}}}_i\in\R^K$ are one-hot vectors in which the position of the sampled topic is set to one during training. For inference, we always choose the topic having the highest probability.

Finally, our topic modeling process is shortly denoted as:
\begin{equation}
    \bm{\alpha}, \bm{\CB}, \bm{r}^{topic} = f_{topic}(d, \Omega),
\end{equation}
where $d$ is the utterance and $\Omega$ is the entire vocabulary. The topicalized representation is the result of full attention on the overall word embedding and thus is less sensitive to local perturbations.

\subsection{Contextualized Representation}
\label{sec:con_rep}
The contextualized representation is obtained in the process of hierarchical emotion dynamics modeling. Let $\{\CD_n\}_{n=1}^N$ be a corpus of $N$ conversations. Each conversation $\CD_n=\{(d^{\lambda_i}_i,y_i)\}_{i=1}^{M_n}$ contains $M_n$ utterance-emotion pair, where $d^{\lambda_i}_i$ is the $i$-th utterance spoken by interlocutor $\lambda_i$, and $y_i$ is the corresponding emotion. The self dependency modeling captures emotional influence within an interlocutor. We use $f_{self}(d^{\lambda_i}_i, c^{\lambda_i}_i)$ to denote the self dependency modeling where $c^{\lambda_i}_i\!=\!\{d^{\lambda_\tau}_\tau|\tau\!\in\![1,i),\lambda_\tau\!=\!\lambda_i\}$ is $\lambda_i$'s historical utterances. The interpersonal dependency modeling, $f_{inter}$, combines the emotional influence across interlocutors. The complete emotion dynamics modeling forms a hierarchical structure, which is
\begin{equation}
    f_{inter}\left(f_{self}(d^{\lambda_1}_1, c^{\lambda_1}_1),f_{self}(d^{\lambda_2}_2, c^{\lambda_2}_2),\cdots\right).
\end{equation}
We drop the superscript $\lambda_i$ for simple in the following part.

Here, we use $\BERT$~\cite{devlin2019bert} as the branches of the hierarchical structure to model $f_{self}$, then the branches are connected to a Transformer~\cite{vaswani2017attention} backbone to model $f_{inter}$. Given an utterance of $L$ words, $d=w_1\,\cdots\,w_{L}$, and its historical context with $L'$ words, $c=\omega_1\,\cdots\,\omega_{L'}$, the input of the \BERT~can be denoted as,
\begin{equation}
    \bm{X} = [\CLS]\,w_1\,\cdots\,w_{L}\,[\SEP]\,\omega_1\,\cdots\,\omega_{L'}\,[\SEP],
\end{equation}
where $[\CLS]$ and $[\SEP]$ are two special words with specific purposes in \BERT. After feeding \BERT~with $\bm{X}$, it outputs the contextualized representations from the last hidden layer at $[\CLS]$ position, which is
\begin{equation}
\label{eq:bert}
    \bm{r}^{context} = f_{self}(d,c)=\BERT(\bm{X}).
\end{equation}
The \BERT~encodes a $M_n$-turn conversation into a sequence of contextualized representations $\bm{R}\!\!\!=\!\!\!\{\bm{r}_i^{context}\}_{i=1}^{M_n}$. By feeding $\bm{R}$ to a \TRM~(TRansFormer), the interpersonal dependency is modeled as,
\begin{equation}
\label{eq:inter}
    \begin{aligned}
    f_{inter}(f_{self}&(d_1,c_1),\cdots,f_{self}(d_i,c_i))&\\
    =&\TRM(\bm{R}, \bm{\eta}_i);(\bm{\eta}_i = \underbrace{11\cdots1}_{i}\underbrace{00\cdots0}_{M_n-i})&.
    \end{aligned}
\end{equation}
Here, we use the last hidden layer of \TRM~at the $i$-th position as the classification representation for the $i$-th turn utterance. $\bm{\eta}_i$ masks the future information.

\begin{table*}[t]
    {\scriptsize The metrics in \textbf{bold} indicate the best values obtained by our models. * refers to a TodKat variant that only performs topic modeling without using the knowledge base.}
    \begin{center}
    \begin{tabular}{p{4.5cm}|c|>{\centering}p{2.2cm}>{\centering}p{2.2cm}>{\centering}p{2.2cm}>{\centering}p{2.2cm}}
    \toprule
    Models &Topic& IEMOCAP & MELD & DailyDialog & EmoryNLP\tabularnewline
    \hline
    \hline    
    HiGRU~\textit{\cite{jiao2019real}}                   &$\times$& 58.28\% & 54.52\% & 51.90\% & 33.54\%\tabularnewline
    DialGCN~\textit{\cite{ghosal-etal-2019-dialoguegcn}} &$\times$& 60.63\% & 56.17\% & 53.73\% & 33.13\%\tabularnewline
    BERT~\textit{\cite{devlin2019bert}}                  &$\times$& -       & 63.49\% & 54.85\% & 41.11\%\tabularnewline
    RoBERTa~\textit{\cite{liu2019roberta}}               &$\times$& -       & 63.75\% & 54.33\% & 40.81\%\tabularnewline
    DialTRM~\textit{\cite{mao2021dialoguetrm}}           &$\times$& 68.41\% & 64.60\% & 56.44\% & 37.13\%\tabularnewline
    CoGBART~\textit{\cite{li2022contrast}}               &$\times$& 64.10\% & 63.66\% & 54.71\% & 37.57\%\tabularnewline
    \hline
    TodKat*~\textit{\cite{zhu2021topic}}                 &$\surd$& 57.38\% & 61.11\% & 53.44\% & 32.62\% \tabularnewline
    VAE (Laplace)                                        &$\surd$& 68.98\% & 65.36\% & 55.11\% & 39.43\% \tabularnewline
    VAE (LogNormal)                                      &$\surd$& 68.95\% & 65.40\% & 56.27\% & 37.80\% \tabularnewline
    VAE (Dirichlet)                                      &$\surd$& 69.19\% & 65.82\% & 56.53\% & 40.35\% \tabularnewline
    VAE (Gamma)                                          &$\surd$& 68.48\% & 64.37\% & 56.34\% & 37.91\% \tabularnewline
    \hline
    \hline
    \textbf{FATRER-multi}   &$\surd$& \textbf{69.69\%} & \textbf{66.28\%} & \textbf{56.57\%} & 39.13\% \tabularnewline
    \textbf{FATRER-single}  &$\surd$& 68.58\% & 65.32\% & 56.46\% & \textbf{41.87\%} \tabularnewline
    \bottomrule
    \end{tabular}
    \caption{Generalization results on four datasets}
    \label{tab:generalization}
    \end{center}
\end{table*}
\subsection{Joint Modeling}
Given the $i$-th utterance $d_i$, corresponding context $c_i$ of the same interlocutor, and a vocabulary set $\Omega$, the process of topic modeling links $d_i$ and $\Omega$ such that the topicalized representation has a global view, and the process of context modeling connects $d_i$ with $c_i$ to enable the contextualized representation with a local view. The regularization from the representation perspective is enforced by simply concatenating the two representations. Specifically, the modeling of emotion dynamics can be reformulated as,
\begin{equation}
    f_{inter}\left(f'_{self}(d_1,c_1,\Omega), f'_{self}(d_2,c_2,\Omega), \cdots\right)
\end{equation}
where $f'_{self}$ is for self-dependency modeling that concatenates output of $f_{topic}$ and the original $f_{self}$ in equation~(\ref{eq:bert}) to form representations with both global and local views. Let $r^{concat}$ be the concatenated representation output by $f'_{self}$, the interpersonal dependency is modeled via feeding the \TRM~with a representation sequence, $\bm{R}\!=\!\{r^{concat}_i\}_{i=1}^{M_n}$. After interpersonal dependency modeling, the global and local views can be fully balanced and mixed through deep layers of multi-head attention within the \TRM. $\{\bm{h}_i\}_{i=1}^{M_n}$ are the output representation sequence for emotional predictions. 

Our joint learning objective contains two terms, a classification loss term $\ell_{CLS}$ and a topic-oriented loss regularization term $\delta_{REC}$. Let $\CY=\{1,\cdots,|\CY|\}$ be the emotion set. The classification loss is to minimize the cross entropy between a sequence of the predicted emotion probabilities $\{\bm{\CP}^{CLS}_i\in\R^{|\CY|}\}_{i=1}^{M_n}$ and a sequence of the ground truth $\{y_i\in\CY\}_{i=1}^{M_n}$:
\begin{eqnarray}
    \bm{\CP}^{CLS}_i\!\!&\!\!=\!\!&\!\!softmax(\bm{h}_i\bm{W}^\top+\bm{b}),\\
    \ell_{CLS}\!\!&\!\!=\!\!&\!\!- \sum_{i=1}^{M_n}{\log{\CP^{CLS}_{i,y_i}}}.
\end{eqnarray}
The topic-oriented loss regularization term is to minimize the KL (Kullback–Leibler) divergence between the document-word posterior distribution \textit{Multinomial}($\bm{\CP}^{REC}_i$) and the observed distribution \textit{Multinomial}($\bm{\gamma}_i)$:
\begin{eqnarray}
    \bm{\CP}^{REC}_i\!\!&\!\!=\!\!&\!\!\bm{\alpha}_i\bm{\beta}^\top,\\
    \delta_{REC}\!\!&\!\!=\!\!&\!\!\frac{1}{M_n} \sum_{i=1}^{M_n}{\sum_{j=1}^V}\gamma_{i,j}\log(\frac{\gamma_{i,j}}{\CP^{REC}_{i,j}}),
\end{eqnarray}
where $\bm{\CP}^{REC}_i$ is our approximated word probabilities conditioned on the $i$-th utterance in a conversation. 

The regularization from the loss perspective is enforced by the product between the two loss terms, which is
\begin{equation}
    \CL= \mu \cdot \ell_{CLS} \cdot \delta_{REC}.
\end{equation}
The values of these two terms serve as the learning rates for each other, which dynamically and interactively adjust the learning weights during the training process. $\mu$ is the global learning rate to control the convergent speed of the model.

\begin{table*}[t]
    {\scriptsize AAA denotes Accuracy After Attack. \textbf{U} denotes merely attacking the target utterance. \textbf{U+C} means attacking both the target utterance and its context. - indicates being incompatible with context perturbations.}
    \begin{center}
    \begin{tabular}{p{3cm}|p{1.8cm}p{1.8cm}|p{1.8cm}p{1.8cm}|p{1.8cm}p{1.8cm}}
    \toprule
         & \multicolumn{2}{c|}{PWWS} & \multicolumn{2}{c|}{TextFooler} & \multicolumn{2}{c}{TextBugger} \\
        Models & {\footnotesize AAA(\textbf{U+C}) } & {\footnotesize AAA(\textbf{U}) } 
               & {\footnotesize AAA(\textbf{U+C}) } & {\footnotesize AAA(\textbf{U}) }
               & {\footnotesize AAA(\textbf{U+C}) } & {\footnotesize AAA(\textbf{U}) }\\ 
        \hline\hline
        CoGBART         & -      & 42.57\% &  -      & 26.73\% &       - & 44.36\%\\
        DialTRM         & 4.10\% & 42.77\% &  7.00\% & 22.38\% &  6.50\% & 38.61\%\\
        TodKat          & -      & 28.32\% &  -      &  8.12\% &       - & 22.97\%\\
        VAE (Laplace)   & 6.50\% & 43.56\% &  4.48\% & 22.18\% &  7.05\% & 39.21\%\\ 
        VAE (LogNormal) & 2.63\% & 40.99\% &  2.57\% & 19.21\% &  6.12\% & 40.40\%\\
        VAE (Dirichlet) & 7.52\% & 41.58\% & 13.05\% & 25.35\% & 15.57\% & 40.79\%\\
        VAE (Gamma)     & 4.49\% & 42.38\% & 13.42\% & 26.73\% & 12.54\% & 43.17\%\\
        \hline\hline
        \textbf{FATRER-multi} & \textbf{15.00\%} & \textbf{46.14\%} & \textbf{19.13\%} & \textbf{31.88\%} & \textbf{22.50\%} & \textbf{44.75\%} \\
        \textbf{FATRER-single} & 14.56\% & 45.74\% & 14.52\% & 23.37\% & 17.52\% & 43.56\% \\
    \bottomrule
    \end{tabular}
    \end{center}
    \caption{Robustness to Adversarial Examples on IEMOCAP}
    \label{tab:robustness_iemocap}
\end{table*}
\begin{table*}[t]
    \begin{center}
    \begin{tabular}{p{3cm}|p{1.8cm}p{1.8cm}|p{1.8cm}p{1.8cm}|p{1.8cm}p{1.8cm}}
    \toprule
         & \multicolumn{2}{c|}{PWWS} & \multicolumn{2}{c|}{TextFooler} & \multicolumn{2}{c}{TextBugger} \\
        Models & {\footnotesize AAA(\textbf{U+C})} & {\footnotesize AAA(\textbf{U}) } 
              & {\footnotesize AAA(\textbf{U+C})} & {\footnotesize AAA(\textbf{U}) }
              & {\footnotesize AAA(\textbf{U+C})} & {\footnotesize AAA(\textbf{U}) }\\ 
        \hline\hline
        CoGBART         & -       & 28.50\% &       - & 11.22\% &       - & 23.96\%\\
        DialTRM         & 11.09\% & 28.91\% &  4.95\% & 11.29\% & 23.38\% & 35.54\%\\
        TodKat          & -       & 28.29\% &       - & 17.03\% &       - & 43.96\%\\
        VAE (Laplace)   & 10.01\% & 27.72\% &  5.54\% & 11.68\% & 20.72\% & 33.47\%\\ 
        VAE (LogNormal) & 12.28\% & 27.72\% &  5.74\% &  9.31\% & 22.82\% & 37.23\%\\
        VAE (Dirichlet) & 13.47\% & 28.71\% &  8.71\% & 14.46\% & 25.32\% & 37.62\%\\
        VAE (Gamma)     & 15.05\% & 27.52\% & 11.29\% & 14.85\% & 24.92\% & 36.44\%\\
        \hline\hline
        \textbf{FATRER-multi} & \textbf{17.23\%} & \textbf{31.68\%} & \textbf{12.28\%} & \textbf{17.43\%} & \textbf{26.28\%} & \textbf{38.61\%} \\
        \textbf{FATRER-single} & 14.56\% & 30.89\% & 8.32\% & 15.05\% & 23.19\% & 37.43\% \\
    \bottomrule
    \end{tabular}
    \end{center}
    \caption{Robustness to Adversarial Examples on MELD}
    \label{tab:robustness_meld}
\end{table*}
\section{Experimental Setup}
\textbf{Datasets.} The benchmark of ERC involves four datasets, including \textit{DailyDialog}~\cite{li2017dailydialog}, \textit{IEMOCAP}~\cite{busso2008iemocap}, \textit{MELD}~\cite{poria2019meld}, and \textit{EmoryNLP}~\cite{zahiri2018emotion}. The first two are dyadic conversational datasets, and the last two comprise multi-party conversations. \textit{IEMOCAP} annotates each utterance with one of 6 emotion labels, and the emotion labels are 7 in the other datasets. We follow the standard split~\cite{zhu2021topic} for training, validation, and testing in the benchmark. The `\textit{neutral}' label is not included in \textit{DailyDialog}'s evaluation to avoid category imbalance.~\cite{ghosal2020cosmic}

\noindent\textbf{Adversarial Attacks.} Three types of adversarial attacks~\footnote{https://github.com/QData/TextAttack}, including \textit{PWWS}~\cite{ren-etal-2019-generating}, \textit{TextFooler}~\cite{jin2019bert}, and \textit{TextBugger}~\cite{li2018textbugger}, are employed for robustness evaluation. \textit{PWWS} offers a word-level attack according to the word saliency and the classification probability. \textit{TextFooler} is reported to have effectiveness to attack pre-training models. \textit{TextBugger} can execute both character-level and word-level attacks. We set all the attackers to perturb up to 25\% of words per input.

\noindent\textbf{Implementation Details.} The single-topic and multi-topic hypotheses yield the \textit{FATRER-single} and \textit{FATRER-multi} versions of our model, respectively. The model described in section~\ref{sec:con_rep} is named as \textit{Baseline}, in which the $\BERT$ is initialized from the $\BERT_{\hbox{Base}}$ and the $\TRM$ is a random initialized, 6-layer, 12-head-attention, and 768-hidden-unit Transformer encoder. Following the spirit of ProdLDA~\cite{srivastava2017autoencoding}, we implement VAE-based topic regularizers underlying different priors. By replacing our topicalized representation with topic variables obtained via variational inference, we have VAE-based variants, including \textit{VAE (Laplace)}, \textit{VAE (LogNormal)}, \textit{VAE (Dirichlet)}, and \textit{VAE (Gamma)}. Based on the same \textit{Baseline}, we can perform a fair comparison between \textit{FATR-based} and \textit{VAE-based} topic regularizers. We use off-the-shelf implementations of \textit{TodKat}\footnote{https://github.com/something678/TodKat} and \textit{CoGBART}\footnote{https://github.com/whatissimondoing/CoG-BART} as well as their trained models to perform robustness evaluations in our experimental settings.

\section{Results, Analysis, and Visualization}
\subsection{Generalization on Benchmark Datasets} 
The generalization results are presented in Table~\ref{tab:generalization}. The listed methods can be divided into two groups according to whether or not latent topics are used. The methods using latent topics can be further categorized as 1) merely pre-trained topics, i.e., \textit{TodKat}, 2) VAE-based joint learned topics, i.e., \textit{VAE (Dirichlet)}, etc., and 3) our \textit{FATR-based} topics. We employ Micro F1 to measure the accuracy of the listed models. The results show that \textit{FATRER-multi} achieves the best performance on the benchmark in terms of Micro F1. Compared with \textit{TodKat} which applies VAE-based topic modeling only in the pre-training stage, we understand the importance of performing joint topic modeling which provides consistent topic guidance for accurate predictions. Compared to adding VAE-based topic regularizers, we can see that our FATR can obtain better generalization results.

\begin{figure}[t]
\centering
    \includegraphics[width=0.48\textwidth]{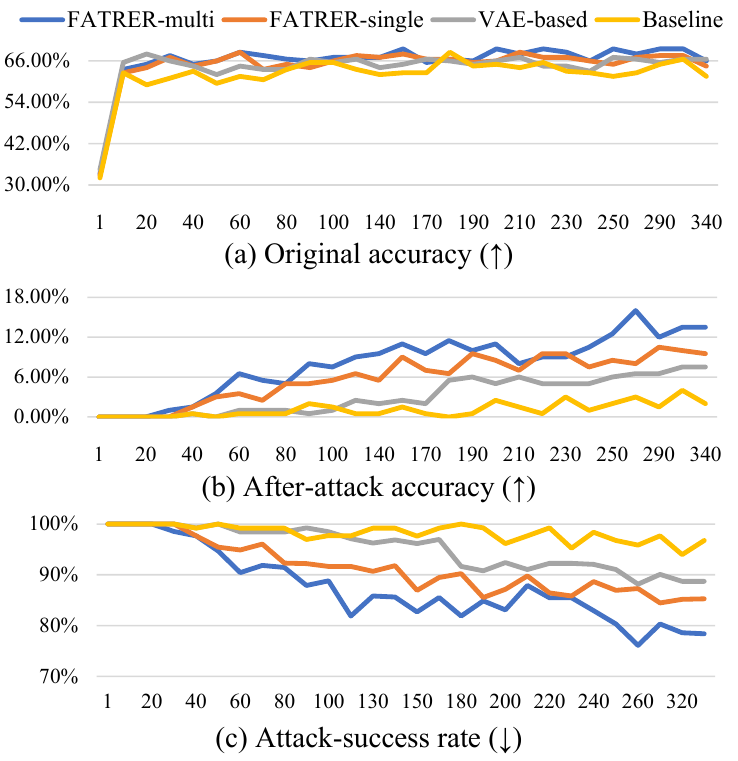}
    \caption{Results over training epochs on IEMOCAP}
    \label{fig:robustness}
\end{figure}

\subsection{Robustness to Adversarial Examples}
Based on the fully converged models, the robustness results are presented in Table~\ref{tab:robustness_iemocap}, \ref{tab:robustness_meld}, and Figure~\ref{fig:robustness}. The evaluation uses AAA (After-Attack Accuracy) and ASR (Attack-Success Rate) as the metrics, where AAA means the accuracy achieved by the target model on the adversarial samples crafted from the test samples. Specifically, AAA(\textbf{U}) denotes AAA of merely attacking target utterance, and AAA(\textbf{U+C}) means AAA of attacking both target utterance and its context. ASR is the proportion of adversarial samples that can successfully change the originally correctly predicted labels.

\begin{figure}[t]
    \centering
    \includegraphics[width=0.48\textwidth]{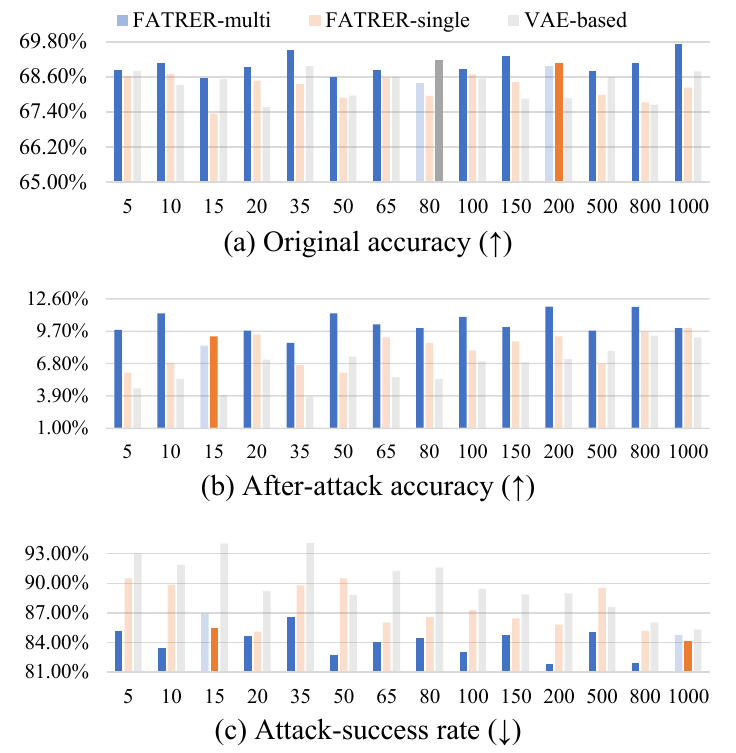}
    \caption{Results over the number of topics on IEMOCAP}
    \label{fig:topic_number}
\end{figure}

Table~\ref{tab:robustness_iemocap} and~\ref{tab:robustness_meld} present the robustness results on IEMOCAP and MELD, respectively. For a fair comparison, the ANQ (Average Number of attack Queries) for each model is tuned to be as equal as possible in this experiment. Our models achieve the best results under the three types of adversarial attacks on the two datasets. We conclude four points from the results:
\begin{itemize}
  \item [1)] 
  Topic-oriented models, e.g., the VAE-based and our \textit{FATRERs}, get better robustness performance, which means the joint modeling of topic and context already provides a certain degree of robustness.    
  \item [2)]
  Our \textit{FATRERs} are superior to VAE-based models, which indicates that our topicalized representation improves robustness beyond the simple use of topic variables.  
  \item [3)]
  The robustness of our \textit{FATRERs} is more evident when the conversational context is disrupted, further demonstrating the benefits of the global view behind our topicalized representations. 
  \item [4)]
  \textit{TodKat} only applies VAE-based topic modeling in the pre-training stage and gets relatively weak performance, which means that without joint topic modeling, it is difficult to guarantee robustness.
\end{itemize}
In a word, the superior robustness of our \textit{FATRERs} mainly comes from the topicalized representation guided by a global view and the full-attention-based joint modeling framework.

In Figure~\ref{fig:robustness}, we track the trend of after-attack accuracy, attack-success rate, and original accuracy (accuracy before the attack) over 340 training epochs. The best performed \textit{VAE (Dirichlet)} is taken as the representative of VAE-based variants. In this experiment, to exert maximum pressure on the models, we let the attacker PWWS determine the number of attack queries and use \textbf{U+C} attack strategy. From the trend in Figure~\ref{fig:robustness}a, we understand that the generalization ability converges quickly within 10 epochs. By contrast, the model robustness in Figure~\ref{fig:robustness}b and~\ref{fig:robustness}c keeps getting better until 300 epochs, which means the model robustness needs a long period of training. From the 200th to 300th training epoch, the ANQ of our models is around 800 which is 1.09 to 1.26 times that of the other models. Higher ANQ means the model is more difficult to be successfully attacked. The results show that our models obtain convincing robustness with greater pressures at every epoch of testing.

\subsection{Analysis}

\textbf{Ablation Study.}  To better understand our models, we yield several variants of our model by removing some key components. Firstly, \textit{FATRER-muti} and \textit{FATRER-single} are two basic variants of our model. \textit{- rep regularizer} indicates removing the topicalized representation and thus is identical to the \textit{Baseline}. \textit{- loss regularizer} means removing the topic-oriented regularization term in the training loss while preserving the attention network for topic modeling, and the topic relative distributions are driven by the classification loss. From the results in Table~\ref{tab:ablation}, we understand that removing either representation or loss regularizer leads to a drop in accuracy and robustness. The model robustness is more sensitive to the removal of the two regularizers than the accuracy.

\begin{table}
    \centering
    \begin{tabular}{p{1.15cm}|p{1.9cm}|c|c|c}
        \toprule
        Datasets&Models & AAA(\textbf{U+C}) & AAA(\textbf{U}) & Micro F1 \\
        \hline
        \hline
        
        \multirow{6}{*}{IEMOCAP}&FATRER-multi & 15.00\% & 46.14\% & 69.69\% \\
        &~-rep regularizer&  4.10\% & 42.77\% & 68.41\% \\
        &~-loss regularizer &  7.04\% &     45.15\% & 69.38\% \\
        \cmidrule{2-5}
        &FATRER-single       & 14.56\% & 45.74\% & 68.58\% \\
        &~-rep regularizer&  4.10\% & 42.77\% & 68.41\% \\
        &~-loss regularizer &  9.03\% &     44.36\% & 68.58\% \\
        \midrule
        \multirow{6}{*}{MELD}&FATRER-multi & 17.23\% & 31.68\% & 66.28\% \\
        &~-rep regularizer& 11.09\% & 28.91\% & 64.60\% \\
        &~-loss regularizer &     14.06\% &     31.09\% &     64.98\% \\
        \cmidrule{2-5}
        &FATRER-single       & 14.65\% & 30.89\% & 65.32\% \\
        &~-rep regularizer& 11.09\% & 28.91\% & 64.60\% \\
        &~-loss regularizer &     13.27\% &     30.69\% &     65.19\% \\
        \bottomrule
    \end{tabular}
    \caption{Ablation study on IEMOCAP and MELD}
    \label{tab:ablation}
\end{table}

\noindent\textbf{Effects of Topic Number.}
We explore the number of topics from 5 to 1000 to analyze their impact on our models. From the results shown in Figure~\ref{fig:topic_number}, we can see that our models have better robustness performance than the \textit{VAE-based} model over the selected numbers of topics. Our model and the VAE-based model achieve the best original accuracy at 80 and 1,000 topics, respectively. Note that, both the number and the content of the adversarial examples are changed for each epoch of testing. Thus, the robustness results shown in Figure~\ref{fig:topic_number}b and~\ref{fig:topic_number}c are based on the average of the scores obtained from the 200th to 300th epoch.

\begin{figure}[ht]
\centering
\begin{subfigure}{0.48\textwidth}
    \includegraphics[width=\textwidth]{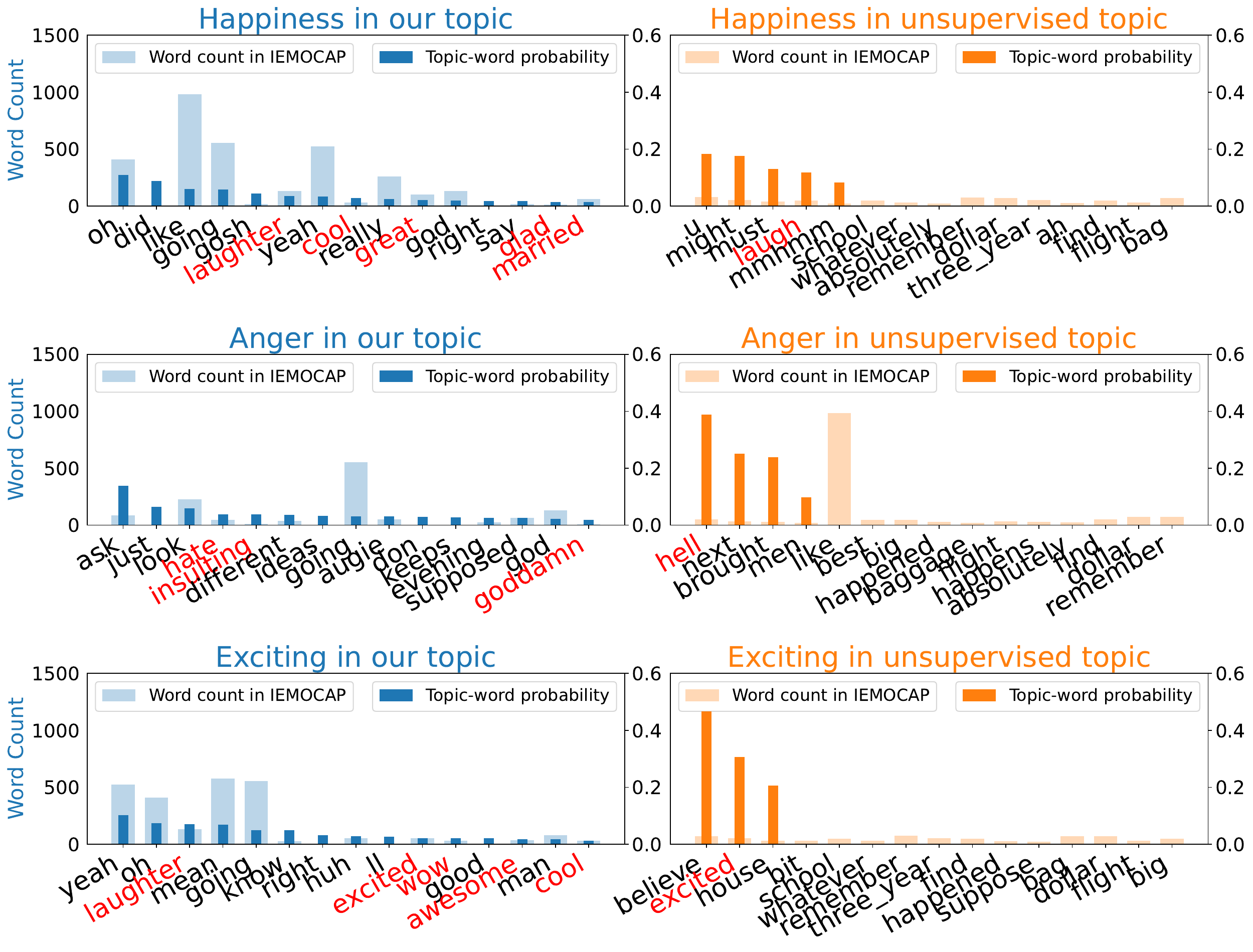}
    \caption{Topic words visualization.}
    \label{fig:visualization1}
\end{subfigure}
\hfill
\begin{subfigure}{0.48\textwidth}
    \includegraphics[width=\textwidth]{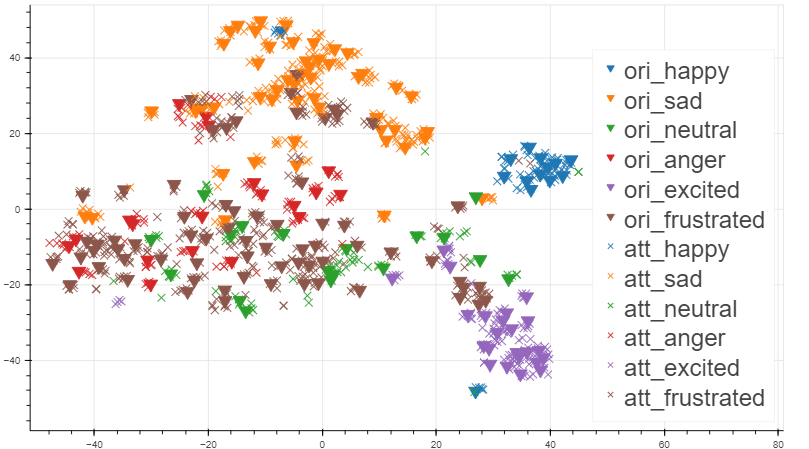}
    \caption{Representation visualization.}
    \label{fig:visualization2}
\end{subfigure}  
\caption{Topic visualization.}
\label{fig:visualization}
\end{figure}

\subsection{Visualization}
\textbf{Topic words visualization.} We evaluate our topics' quality by comparison with unsupervised topics learned from LDA~\cite{blei2003latent}. By inspecting the top 15 keywords, we depict three topics that are most relative to a kind of emotion in Figure~\ref{fig:visualization1}. We mark the emotional words in red. Benefiting from the supervision of emotion labels, our topic model can discover more emotional words than LDA. Such emotion-related topics can improve accuracy. The top words of our topics have perceptible probabilities while LDA's topics have less than 5 valid keywords. Both frequent and rare words are learned in our topics. Such wide-range and emotion-related global views can guarantee robustness.  

\noindent\textbf{Representation visualization.} To give an intuitive understanding of the adversarial attacks, we plot representations of the \textbf{ori}ginal and the \textbf{att}acked samples output from \textit{FATRER-multi} in Figure~\ref{fig:visualization2}. We randomly choose 5 out of 800+ attacked samples crafted from each original sample. As shown in the figure, the representations of the attacked samples wrap tightly around the original samples. By the way, we notice that our model still has room for the discrimination of neutral, angry, and frustrated emotions, which is a clue for improving our model in the future.

\section{Conclusion}
We propose a topic-augmented conversational emotion recognizer, namely FATRER, for the task of ERC. To cope with the impact of local perturbations, the discovered topic is enabled with an emotion-related global view based on a full-attention framework. By jointly performing topic and context modeling, our full-attention topic regularizer augmented models can achieve accurate
and robust conversational emotion recognition. Experimental results on standard benchmark datasets and adversarial examples have shown the generalization and robustness of our models. Ablation studies are conducted to help understand the effectiveness of our topic regularizers. We also provide visualization to give an intuitive explanation of our models.

\bibliography{ecai}
\end{document}